\pdfoutput=1

\documentclass[11pt]{article}

\usepackage[preprint]{acl}

\usepackage{times}
\usepackage{latexsym}

\usepackage[T1]{fontenc}

\usepackage[utf8]{inputenc}

\usepackage{siunitx}

\usepackage{microtype}

\usepackage{inconsolata}

\usepackage{graphicx}
\usepackage[x11names,table]{xcolor}

\usepackage{amsmath}
\usepackage{amssymb}
\usepackage{bbm}
\usepackage{mathtools}
\usepackage{booktabs}
\usepackage{nicematrix}
\usepackage{multirow}
\usepackage{arydshln}
\usepackage{adjustbox}
\usepackage{algorithm}
\usepackage[indLines=true,noEnd=false]{algpseudocodex}
\usepackage{svg}
\usepackage{bm}
\usepackage{float}
\usepackage{placeins}

\newcommand{\regmark}{${}^{\text{\textregistered}}$}

\newcommand{\colorGray}{gray!20}

\algrenewcommand\algorithmicindent{0.75em}

\usepackage{cleveref}
\crefformat{section}{#2Section #1#3}
\crefformat{subsection}{#2Section #1#3}
\crefname{equation}{Equation}{Equation}
\crefname{figure}{Figure}{Figure}
\crefname{table}{Table}{Table}
\crefname{appendix}{Appendix}{Appendix}
\crefname{line}{Line}{Line}
\crefname{algorithm}{Algorithm}{Algorithm}

\DeclareMathOperator{\argmax}{argmax}
\DeclareMathOperator{\argmin}{argmin}
\DeclareMathOperator{\E}{\mathbb{E}}
\newcommand{\observe}[1]{\mathrm{Obs}({#1})}
\newcommand{\R}{\mathbb{R}}
\newcommand{\N}{\mathbb{N}}

\newcommand{\defFunc}[3]{{#1}\colon{#2}\to{#3}}

\newcommand{\setHyps}{\mathcal{Y}}

\newcommand{\numHyps}{N}
\newcommand{\numPseudoRefs}{M}
\newcommand{\score}{O}
\newcommand{\scoreMatrix}{\mathbf{\score}}
\newcommand{\scoreMatrixIncomplete}{\tilde{\scoreMatrix}}
\newcommand{\utility}{u}
\newcommand{\reduction}{r}
\newcommand{\loss}{\mathcal{L}}
\newcommand{\lossMF}{\loss_\mathrm{MF}}
\newcommand{\lossAC}{\loss_\mathrm{AC}}

\title{Agreement-Constrained Probabilistic Minimum Bayes Risk Decoding}

\author{
  \textbf{Koki Natsumi}${}^\dagger$, ~~~
  \textbf{Hiroyuki Deguchi}${}^\ddagger$, \\ 
  \textbf{Yusuke Sakai}${}^\dagger$, ~~~
  \textbf{Hidetaka Kamigaito}${}^\dagger$, ~~~
  \textbf{Taro Watanabe}${}^\dagger$
\\
  ${}^\dagger$Nara Institute of Science and Technology (NAIST) ~~~
  ${}^\ddagger$NTT, Inc.
\\
  \texttt{natsumi.koki.ng5@naist.ac.jp, hiroyuki.deguchi@ntt.com} \\
  \texttt{\{sakai.yusuke.sr9, kamigaito.h, taro\}@is.naist.jp}
}

\begin{document}
\maketitle
\begin{abstract}
Minimum Bayes risk (MBR) decoding generates high-quality translations by maximizing the expected utility of output candidates, but it evaluates all pairwise scores over the candidate set; hence, it takes quadratic time with respect to the number of candidates.
To reduce the number of utility function calls, probabilistic MBR (PMBR) decoding partially evaluates quality scores using sampled pairs of candidates and completes the missing scores with a matrix completion algorithm.
Nevertheless, it degrades the translation quality as the number of utility function calls is reduced.
Therefore, to improve the trade-off between quality and cost, we propose agreement-constrained PMBR (AC-PMBR) decoding, which leverages a knowledge distilled model to guide the completion of the score matrix.
Our AC-PMBR decoding improved approximation errors of matrix completion by up to 3 times and achieved higher translation quality compared with PMBR decoding at a comparable computational cost on the WMT'23 En$\leftrightarrow$De translation tasks.
\end{abstract}

\section{Introduction}
Maximum a posteriori (MAP) decoding, which finds the most probable candidate, is the standard inference strategy in translation tasks, while such high‑probability translations do not always align with human assessment~\cite{koehn-knowles-2017-six,eikema-aziz-2020-map}.
To overcome the limitation, minimum Bayes risk (MBR) decoding selects a high-quality translation rather than a high-probability one by maximizing expected utility~~\cite{goel-and-byrne-2000-minimum,kumar-byrne-2004-minimum}.
For estimating expected utility, it calculates the utility score matrix, evaluating all candidates against multiple pseudo‑references, which are sample translations drawn from the output distribution.
Thus, it requires utility function calls proportional to the square of the number of candidates and is computationally expensive, especially when using neural metrics that highly correlate with human assessment, e.g., BLEURT~\cite{sellam-etal-2020-bleurt}.

Recent studies improve the efficiency of MBR decoding~\cite{cheng-vlachos-2023-faster,jinnai-ariu-2024-hyperparameter,deguchi-etal-2024-centroid,vamvas-sennrich-2024-linear,trabelsi-etal-2024-efficient}. Among them, probabilistic MBR (PMBR) decoding~\cite{trabelsi-etal-2024-efficient} drastically reduces the number of utility function calls by completing the score matrix using partially observed scores.
Nevertheless, as the number of utility function calls is reduced, the approximation error of matrix completion increases, and the translation quality deteriorates.
That is, there exists a trade-off between completion accuracy and computational cost.

To relax this trade-off, we propose \emph{agreement-constrained PMBR (AC-PMBR)} decoding, which facilitates score matrix completion by leveraging a knowledge distilled metric.
Our agreement constraint minimizes the difference between the target and distilled low‑rank matrices, thereby reducing the approximation error in matrix completion.

Experiments demonstrated that our AC-PMBR decoding improved the matrix completion accuracy by up to 3 times in mean squared error (MSE) against the full score matrix and achieved 
higher translation quality compared with PMBR decoding at comparable costs in the WMT'23 En$\leftrightarrow$De translation tasks~\cite{kocmi-etal-2023-findings}.

\begin{figure}[t]
 \centering
 \includesvg[width=1.0 \linewidth]{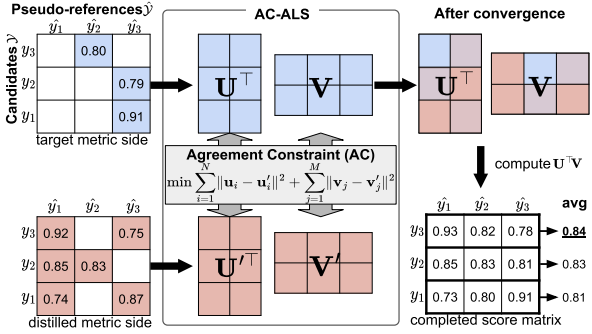}
 \caption{Overview of our proposed Agreement-Constrained PMBR (AC-PMBR) decoding.}
 \label{fig:AC-PMBR}
\end{figure}

\section{Background}
\paragraph{MBR decoding}
MBR decoding finds higher-quality translations than widely used MAP decoding, such as $N$-best beam search, by maximizing the expected utility of output candidates~\cite{kumar-byrne-2004-minimum, eikema-aziz-2020-map}.
Let $\mathcal{T}$ be all possible translations.
The goal of MBR decoding is to find the translation that maximizes the expected utility, i.e.,
$\argmax_{y \in \mathcal{T}} \E_{\hat{y} \sim \Pr(y|x)} [u(y, \hat{y})]$,
where $x$ is an input text, $\Pr(y|x)$ is the true translation probability, and $\defFunc{\utility}{\mathcal{T} \times \mathcal{T}}{\R}$ denotes a utility function that satisfies $y \succeq y' \iff \utility(y, \hat{y}) \geq \utility(y', \hat{y})$ under the preference relation $\succeq$.
Since enumerating all $y \in \mathcal{T}$ is infeasible and calculating the true probability $\Pr$ is unknown, MBR decoding estimates the expected utility using sample translations drawn from the model output probability, called pseudo-references $\hat{\mathcal{Y}} \coloneqq \{\hat{y}_1, \dots, \hat{y}_M\} \subset \mathcal{T}$, and selects the translation from a candidate set $\mathcal{Y} \coloneqq \{y_1, \dots, y_N\} \subset \mathcal{T}$.
The expected utility is typically estimated by the Monte Carlo method~\citep{eikema-aziz-2022-sampling} with a score matrix $\scoreMatrix \coloneqq [\score_{ij} = \utility(y_i, \hat{y}_j)] \in \R^{\numHyps \times \numPseudoRefs}$, and then, the best candidate is selected, i.e., $y_\mathrm{MBR} \coloneqq \argmax_{y_i \in \setHyps} \frac{1}{\numPseudoRefs} \sum_{j=1}^{\numPseudoRefs} \score_{ij}$.

MBR decoding generates high-quality translations, while its time complexity is $\mathcal{O}(\numHyps \numPseudoRefs)$.
Recent studies often employ $N \ge 1{,}000$~\citep{freitag-etal-2023-epsilon}, making it extremely slow.

\paragraph{PMBR decoding}
Probabilistic MBR (PMBR) decoding accelerates MBR decoding by reducing the number of utility function calls~\citep{trabelsi-etal-2024-efficient}.
It does not evaluate scores for all $O_{ij}$; instead, it partially evaluates only sampled pairs of hypotheses and pseudo-references.
The other missing scores are completed using a low-rank matrix factorization from a partially observed score matrix $\scoreMatrixIncomplete\in\R^{\numHyps\times \numPseudoRefs}$.
Specifically, the incomplete matrix $\scoreMatrixIncomplete$ is approximated by the matrix multiplication of two $d$-dimensional low-rank matrices $\mathbf{U}\in\R^{d \times \numHyps}$ and $\mathbf{V}\in\R^{d \times \numPseudoRefs}$, i.e., $\scoreMatrixIncomplete \approx \mathbf{U}^\top \mathbf{V}$.
Here, $\mathbf{u}_i, \mathbf{v}_j\in\mathbb{R}^d$ are $d$-dimensional column vectors, and $\mathbf{U}=[\mathbf{u}_1; \dots; \mathbf{u}_N]$ and $\mathbf{V}=[\mathbf{v}_1; \dots; \mathbf{v}_M]$ stack the rank reduced vectors for the row and column directions of $\tilde{\mathbf{O}}$, respectively.  
Let $\observe{\scoreMatrixIncomplete} \coloneqq\{(i,j) \mid \tilde{O}_{ij}\text{ is observed}\}$ be the set of observed indices in $\scoreMatrixIncomplete$.
We obtain $\mathbf{U}$ and $\mathbf{V}$ that minimize the following objective:
\begin{align}
    \lossMF(\mathbf{U}, \mathbf{V}&; \scoreMatrixIncomplete) =
        \sum\nolimits_{(i,j)\in \observe{\scoreMatrixIncomplete}} ( \tilde{\score}_{ij}-\mathbf{u}_i^\top \mathbf{v}_j )^2 \nonumber \\
        + &\lambda ( \sum\nolimits_{i=1}^{N} \lVert \mathbf{u}_i\rVert^2 + \sum\nolimits_{j=1}^{M} \lVert\mathbf{v}_j\rVert^2 ),
\label{eq:pmbr}
\end{align}
where $\lambda \in \R_+$ is a weight of the regularization term.
This optimization is solved by the alternating least-squares (ALS) algorithm~\cite{Zachariah2012AlternatingLF},
and the complete score matrix is obtained by $\mathbf{U}^{\top}\mathbf{V}$.
PMBR decoding successfully reduces the number of utility function calls.
Nevertheless, there is still a cost-quality trade‑off, because further reductions in utility function calls significantly degrade approximation accuracy.

\section{Agreement-Constrained PMBR decoding}
We propose agreement‑constrained PMBR (AC‑PMBR) decoding, which alleviates the PMBR decoding cost–quality trade‑off without increasing total cost by reallocating a fixed evaluation budget.
Instead of adding the distilled metric on top of PMBR decoding, AC‑PMBR decoding reduces target metric calls and spends the saved budget on many distilled metric calls. Thereby enabling more total utility function calls at the same computational cost as PMBR decoding and yielding higher matrix‑completion accuracy of the MBR's score matrix. AC‑PMBR decoding proceeds in two steps: (1) score matrix construction, and (2) agreement‑constrained matrix completion, as illustrated in Figure~\ref{fig:AC-PMBR}.

\paragraph{Score matrix construction}
We compute the score matrices, $\scoreMatrixIncomplete \in \R^{\numHyps\times\numPseudoRefs}$ and $\scoreMatrixIncomplete' \in \R^{\numHyps\times\numPseudoRefs}$, with the target and its distilled metrics, respectively.
Hereafter, we denote a prime $'$ for the distilled metric side.
Let $\reduction$ and $\reduction'$ denote the reduction rates; we observe only a $1/\reduction$ and $1/\reduction'$ fraction of the $N\times M$ entries in $\scoreMatrixIncomplete$ and $\scoreMatrixIncomplete'$, respectively.
The time complexity of evaluating the partially observed samples of hypotheses and pseudo‑references is
$\mathcal{O}(\frac{\numHyps \numPseudoRefs}{\reduction})$.
As the reduction rate $\reduction$ or $\reduction'$ increases, the number of observed samples decreases.
To alleviate the cost--quality trade‑off, we set $\reduction > \reduction'$, i.e., we call a distilled metric more frequently than an expensive target metric for denser guidance at almost the same cost.

\paragraph{Agreement-constrained matrix completion}
We factorize the matrices $\scoreMatrixIncomplete$ and $\scoreMatrixIncomplete'$ with the alternating least squares (ALS) algorithm~\citep{Zachariah2012AlternatingLF}, i.e., we minimize $\lossMF(\mathbf{U}, \mathbf{V}, \scoreMatrixIncomplete)$ and $\lossMF(\mathbf{U}', \mathbf{V}', \scoreMatrixIncomplete')$ with our proposed agreement constraint.
The constraint encourages the rank reduced representation on the target metric to be closer to that of its knowledge distilled metric:
\begin{align}
\label{eq:loss:agreement}
    &\lossAC(\mathbf{U}, \mathbf{V}, \mathbf{U}', \mathbf{V}') \nonumber \\
    &= \sum\nolimits_{i=1}^{N}\lVert\mathbf{u}_i-\mathbf{u}'_i\rVert^2
     +\sum\nolimits_{j=1}^{M}\lVert\mathbf{v}_j-\mathbf{v}'_j\rVert^2.
\end{align}
Formally, our AC-PMBR decoding minimizes the following objective:
\begin{align}
\label{eq:loss:overall}
    \argmin_{\mathbf{U}, \mathbf{V}}
    &\lossMF(
        \mathbf{U}, \mathbf{V}; \scoreMatrixIncomplete
    ) + \lossMF(
        \mathbf{U}', \mathbf{V}'; \scoreMatrixIncomplete'
    ) \nonumber \\
    &+ \gamma\lossAC(\mathbf{U}, \mathbf{V}, \mathbf{U}', \mathbf{V}'),
\end{align}
where $\gamma \in \R_+$ controls the strength of the agreement constraint.
This constrained optimization problem can be solved by extending the ALS algorithm, as shown in Algorithm~\ref{alg:ac-pmbr}; a detailed derivation is provided in Appendix~\ref{sec:appendix}.
The rank reduced representations of the distilled metric side are first updated, as shown in \cref{alg:acals:distill-u} and \ref{alg:acals:distill-v}, so that those of the target metric side are not affected by the unupdated matrices that do not have meaningful information.
Now, we complete the incomplete score matrix by multiplying matrices $\mathbf{U}$ and $\mathbf{V}$, calculated in Equation \ref{eq:loss:overall}, i.e., we use $\mathbf{U}^\top \mathbf{V}$ as the completed score matrix and estimate the expected utility in the same way as MBR decoding.

\begin{algorithm}[t]
\footnotesize
\caption{Agreement-constrained ALS}
\label{alg:ac-pmbr}
\begin{algorithmic}[1]
\Require Regularization weight $\lambda \in \R_+$, agreement weight $\gamma \in \R_+$, rank $d \in \N$, and identity matrix $\mathbf{I} \in \R^{d \times d}$
\Ensure $\mathbf{U} \in \R^{d \times N}$ and $\mathbf{V} \in \R^{d \times M}$
\Repeat
\State Initialize $\mathbf{U}, \mathbf{U}' \in \R^{d \times N}$ and $\mathbf{V}, \mathbf{V}' \in \R^{d \times M}$
\For{$i = 1 \dots N$}
    \State $\mathbf{M}' = \mathrm{diag}(\mathbbm{1}_{\observe{\scoreMatrixIncomplete'}} [(i, 1)], \ldots, \mathbbm{1}_{\observe{\scoreMatrixIncomplete'}} [(i, M)])$
    \State $\mathbf{M} = \mathrm{diag}(\mathbbm{1}_{\observe{\scoreMatrixIncomplete}} [(i, 1)], \ldots, \mathbbm{1}_{\observe{\scoreMatrixIncomplete}} [(i, M)])$
    \State $\mathbf{u}'_i = (
    \mathbf{V}' \mathbf{M}' {\mathbf{V}'}^\top + (\lambda + \gamma) \mathbf{I})^{-1}(
    \mathbf{V}' \mathbf{M}'\scoreMatrixIncomplete'_{i*}
    \label{alg:acals:distill-u}
    + \gamma \mathbf{u}_i)$
    \State $\mathbf{u}_i = (
    \mathbf{V} \mathbf{M} \mathbf{V}^\top + (\lambda + \gamma) \mathbf{I})^{-1}(
    \mathbf{V} \mathbf{M}\scoreMatrixIncomplete_{i*}
    + \gamma \mathbf{u}'_i)$
\EndFor

\For{$j = 1 \dots M$}
    \State $\mathbf{N}' = \mathrm{diag}(\mathbbm{1}_{\observe{\scoreMatrixIncomplete'}} [(1, j)], \ldots, \mathbbm{1}_{\observe{\scoreMatrixIncomplete'}} [(N, j)])$
    \State $\mathbf{N} = \mathrm{diag}(\mathbbm{1}_{\observe{\scoreMatrixIncomplete}} [(1, j)], \ldots, \mathbbm{1}_{\observe{\scoreMatrixIncomplete}} [(N, j)])$
    \State $\mathbf{v}'_j = (
    \mathbf{U}' \mathbf{N}' {\mathbf{U}'}^\top + (\lambda + \gamma) \mathbf{I})^{-1}(
    \mathbf{U}' \mathbf{N}'\scoreMatrixIncomplete'_{*j}
    \label{alg:acals:distill-v}
    + \gamma \mathbf{v}_j)$
    \State $\mathbf{v}_j = (
    \mathbf{U} \mathbf{N} \mathbf{U}^\top + (\lambda + \gamma) \mathbf{I})^{-1}(
    \mathbf{U} \mathbf{N}\scoreMatrixIncomplete_{*j}
    + \gamma \mathbf{v}'_j)$
\EndFor
\Until convergence
\State \Return $\mathbf{U}, \mathbf{V}$
\label{ALS}
\end{algorithmic}
\end{algorithm}

\section{Experimental Settings}
\begin{table*}[t]
  \centering
  \small
  \setlength{\tabcolsep}{3pt}
  \begin{NiceTabular}{@{}l r rrrrrr rrrrrr@{}}[colortbl-like]
    \toprule
    & &
    \multicolumn{6}{c}{En$\rightarrow$De} &
    \multicolumn{6}{c}{De$\rightarrow$En} \\
    \cmidrule(lr){3-8}\cmidrule(l){9-14}
    Decoding & Dist.&
    {BLRT$\uparrow$} & {XCT$\uparrow$} & {BLEU$\uparrow$} & {chrF$\uparrow$} & {MX$\downarrow$} & {MSE$\downarrow$} &
    {BLRT$\uparrow$} & {XCT$\uparrow$} & {BLEU$\uparrow$} & {chrF$\uparrow$} & {MX$\downarrow$} & {MSE$\downarrow$} \\
    \midrule %
    MAP & {--}& {45.27} & 59.61 & 10.74 & 30.38 & 12.94 & {--}
                          & {56.27} & {65.79} & {16.56} & {37.39} & {11.68} & {--} \\
    MBR & {--} & {57.42} & 67.83 & 18.97 & 46.19 & 8.87 & {--}
                          & {65.19} & {77.49} & {23.86} & {50.99} & {8.46} & {--} \\
     & {D3} & {46.68} & {57.20} & {18.12} & {46.69} & {10.79} & {10.38}
                          & {60.65} & {70.67} & {23.37} & {51.89} & {9.66} & {8.53} \\
     & {D6} & {48.89} & {59.54} & {18.31} & {46.49} & {10.33} & {9.43}
                          & {61.80} & {73.42} & {24.28} & {51.37} & {9.38} & {6.97} \\
     & {D12} & {51.33} & {63.54} & {17.73} & {44.53} & {9.79} & {9.58}
                          & {62.21} & {74.28} & {23.92} & {50.55} & {9.45} & {6.53} \\
    \midrule
    \multicolumn{14}{@{}l@{}}{\rowcolor{\colorGray}\textbf{Reduction rate: Low}~~~~ PMBR: $r=16$, ~AC-PMBR: $r=32$
    } \\
    PMBR & {--} & {57.19} & {67.79} & {19.05} & {46.21} & {8.81} & \textbf{3.04}
                          & {64.87} & {77.27} & {23.47} & {50.72} & \textbf{8.46} & \textbf{2.54} \\
    AC-PMBR & {D3} & {57.01} & {67.57} & \textbf{19.19} & \textbf{46.60} & {8.80} & {3.23}
                          & {64.87} & {77.01} & {23.82} & {51.08} & {8.51} & {2.80} \\
                        & {D6} & {57.26} & \textbf{68.00} & 19.14 & 46.27 & \textbf{8.74} & {3.12}
                          & {64.87} & {77.26} & {23.69} & {50.95} & {8.51} & {2.68} \\
                        & {D12} & \textbf{57.29} & 67.94 & 19.09 & 46.23 & {8.77} & {3.10}
                          & \textbf{64.95} & \textbf{77.47} & \textbf{24.05} & \textbf{51.10} & \textbf{8.46} & {2.65} \\
    \midrule
    \multicolumn{14}{@{}l@{}}{\rowcolor{\colorGray}\textbf{Reduction rate: High}~~~ PMBR: $r=512$, ~AC-PMBR: $r=1{,}024$
    } \\
    PMBR & {--} & {50.42} & {60.92} & {16.76} & {44.05} & {10.80} & {26.06}
                          & {60.02} & {70.74} & {21.34} & {48.69} & {10.37} & {33.80} \\
    AC-PMBR & {D3} & {50.49} & {60.56} & {18.15} & \textbf{46.17} & 10.42 & \textbf{10.29}
                          & {61.08} & {71.98} & \textbf{22.96} & \textbf{50.50} & \textbf{9.71} & \textbf{11.89} \\
                          & {D6} & {51.21} & {61.92} & \textbf{18.23} & {45.75} & \textbf{10.12} & {11.16}
                          & {61.02} & {72.10} & {22.83} & {49.94} & {9.90} & {12.57} \\
                          & {D12} & \textbf{51.81} & \textbf{63.41} & 17.54 & 44.48 & 
                          \textbf{10.12} & {15.78}
                          & \textbf{61.26} & \textbf{72.66} & {22.31} & {49.12} & {9.97} & {16.80} \\
    \midrule
    Oracle & {--} & {57.43} & 69.10 & 18.60 & 44.54 & 8.52 & {--}
                          & {70.93} & {79.28} & {26.14} & {52.51} & {7.21} & {--} \\
    \bottomrule
  \end{NiceTabular}
  \caption{Results of WMT'23 En$\leftrightarrow$De.
  BLEURT is abbreviated as BLRT, XCOMET as XCT, and MetricX as MX. ``Dist.'' indicates distilled metrics.
  The best scores within each reduction rate setting are highlighted in \textbf{bold}.}
  \label{tab:score_wmt23}
\end{table*}

We provide the experimental settings below; further details are available in Appendix~\ref{sec:detailed-metrics}.

\paragraph{Evaluation}
We conduct experiments on the WMT'23 En$\leftrightarrow$De translation tasks~\cite{kocmi-etal-2023-findings}.
We evaluate translation quality using BLEURT~\cite{sellam-etal-2020-bleurt}, XCOMET~\cite{guerreiro-etal-2024-xcomet}, BLEU~\cite{papineni-etal-2002-bleu}, chrF~\cite{popovic-2015-chrf}, and MetricX~\cite{juraska-etal-2024-metricx}.
To assess the performance of each method in completing the score matrix, we compute the mean squared error (MSE) between the matrix completed by ALS or Agreement-constrained ALS and the ground-truth matrix. 

\paragraph{Methods}
We compare AC-PMBR decoding with PMBR decoding, and also evaluate MAP and MBR decoding.
We also evaluate the upper bound of translation quality in candidate sets by selecting translations that maximize the target metric using references (Oracle).
Following~\citet{deguchi-etal-2024-mbrs}, we sampled 1{,}024 translation candidates via $\varepsilon$-sampling with $\varepsilon=0.02$~\cite{freitag-etal-2023-epsilon} from M2M100~\cite{DBLP:journals/corr/abs-2010-11125} and used the same set as pseudo-references.

\paragraph{Utility function}
We employ BLEURT-20 as the target metric, along with its three distilled versions, BLEURT-20-\{D3, D6, D12\}~\cite{pu-etal-2021-learning}, as the distilled metrics.
To ensure comparable computational costs between PMBR and AC-PMBR decoding, we choose the reduction rates $r$ and $r'$ in two settings: a low-reduction (\textbf{Low}) setting, where we use $r=16$ for PMBR and $r=32$ for AC-PMBR decoding, with $r' = \{2, 4, 8\}$ for BLEURT-20-\{D3, D6, D12\}, respectively; and a high-reduction (\textbf{High}) setting, where we use $r=512$ for PMBR and $r=1{,}024$ for AC-PMBR decoding, with $r' = \{64, 128, 256\}$, respectively.

\paragraph{Hyperparameter}
The agreement weight $\gamma$ and regularization weight $\lambda$ were tuned on the WMT'22 En$\rightarrow$De translation task~\cite{kocmi-etal-2022-findings} by minimizing the MSE of the score matrices.
ALS and agreement-constrained ALS were run for up to 30 iterations or until the loss difference fell below $10^{-4}$. 
For both PMBR and AC-PMBR decoding, we fixed the rank $d=8$ and $\lambda=0.1$.
In AC-PMBR decoding, we also tuned $\gamma \in \{0.1, \ldots, 1.0\}$ for each reduction rate $r$, and used the optimal values $\gamma=0.1$ for $r=32$ and $\gamma=1.0$ for $r=1{,}024$.

\begin{figure}[t]
    \centering
    \includesvg[width=1.0\linewidth]{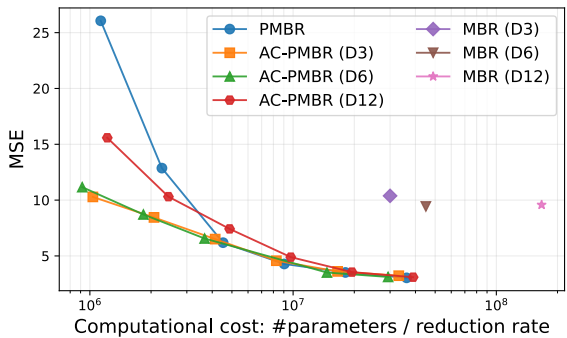}
    \caption{MSE of score matrices when varying computational costs in WMT'23 En$\rightarrow$De. \#parameters refers to the number of parameters in the evaluation metric model, serving as an indicator of model scale, while the computational cost is described by Equation~\ref{eq:costPMBR} and~\ref{eq:costACPMBR}.}
    \label{fig:MSE}
\end{figure}

\begin{figure}[t]
    \centering
    \includesvg[width=1.0\linewidth]{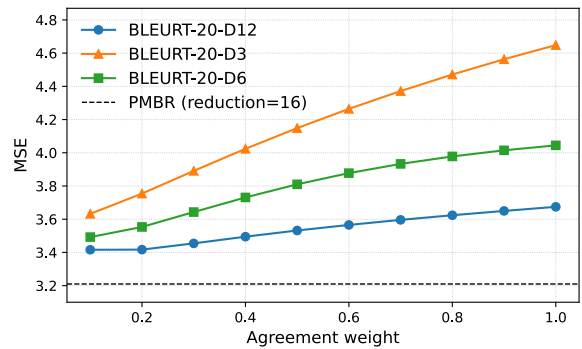}
    \caption{Agreement weight tuning in the low reduction rate setting on WMT'22 En$\rightarrow$De.}

    \label{fig:agreement32}
\end{figure}

\begin{figure}[t]
    \centering
    \includesvg[width=1.0\linewidth]{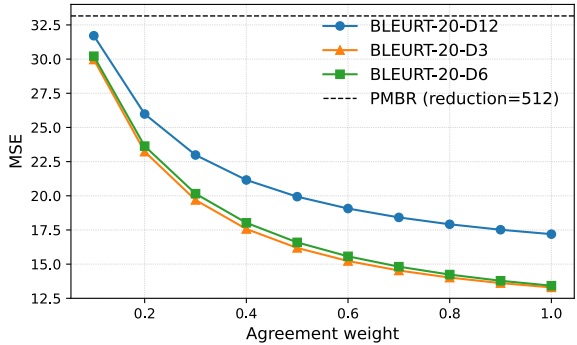}
    \caption{Agreement weight tuning in the high-reduction setting on WMT'22 En$\rightarrow$De.}
    \label{fig:agreement1024}
\end{figure}

\section{Experimental Results and Discussions}

Table~\ref{tab:score_wmt23} shows the main results.

\subsection{Translation quality}
Under the low-reduction setting, AC-PMBR decoding retains an advantage over baseline PMBR in both translation directions. It delivers gains of up to 0.6\% BLEU and 0.4\% chrF on surface-form metrics, and up to 0.2\% on the semantic metric XCOMET.
Under the high-reduction setting, PMBR’s quality fell sharply, whereas AC-PMBR decoding curbed that decline in both directions and stayed ahead on every metric.
Across the two directions, AC-PMBR decoding delivered roughly 2.5\% higher XCOMET and up to 2\% higher chrF, clearly outperforming baseline PMBR decoding in the most demanding scenario.
The results of the significance tests are reported in Appendix~\ref{sec:significance-tests}.

\subsection{Matrix completion accuracy}
In the low-reduction setting, AC-PMBR decoding achieved performance comparable to PMBR decoding in MSE evaluation. In addition, the MSE was significantly improved in the high-reduction setting, with up to a 3 times improvement. This setting has the highest number of utility function calls, and therefore, it is more effective for MSE to evaluate a large number of pairs with low accuracy than to calculate a small number of pairs with a high accuracy model at low computational cost. Figure~\ref{fig:MSE} also shows that AC-PMBR decoding suppresses MSE, which worsens as computational cost decreases.
Moreover, AC-PMBR decoding achieved higher accuracy at a lower cost than MBR decoding with distilled metrics.
This suggests that distilled metrics are effective when used to assist matrix completion via our agreement constraint, rather than being used for the utility function.

\subsection{Effect of Reduction Rate on Translation}
Table~\ref{tab:score_wmt23} shows that with the proposed AC-PMBR decoding, the decline in the approximation accuracy of the score matrix is more gradual compared to conventional PMBR decoding, especially as the reduction rate $r$ increases. This suggests that AC-PMBR decoding can perform matrix completion more accurately than PMBR decoding at a comparable computational cost. Furthermore, as shown in Figures~\ref{fig:agreement32} and ~\ref{fig:agreement1024}, when the reduction rate $r$ is small, a low agreement weight is sufficient because the information from the target model alone is adequate for matrix completion. Conversely, at high reduction rates where the target model's information is insufficient, the information from the knowledge distilled model effectively contributes to the completion process. The fact that its translation quality surpasses that of MBR decoding using only the knowledge distilled model also indicates that our method effectively incorporates information from the target model, even under high reduction rates.

\subsection{Approximation Accuracy of the Score Matrix}
As shown in Figure~\ref{fig:MSE}, the MSE evaluation of the score-matrix approximation clearly demonstrates the robustness of the proposed AC‑PMBR decoding, especially at high reduction rates, where conventional PMBR decoding collapses.
Under high reduction, the number of observed scores becomes critically small, turning matrix completion in PMBR decoding into an ill-posed problem and causing a sharp drop in approximation accuracy. This is reflected in a dramatic increase in MSE and a notable decline in translation quality.
In contrast, AC‑PMBR decoding maintains significantly lower MSE under the same conditions. This robustness is attributed to the information from the dense score matrix provided by the knowledge-distilled model. We presume that this matrix, which captures the general distribution of the true score matrix, acts as a guide that prevents the approximation from failing.
Given that the performance of both methods is comparable at low reduction rates, where observed scores are relatively abundant, AC‑PMBR decoding is expected to perform effectively in more difficult, information-scarce situations.

\section{Conclusion}
In this study, we proposed AC-PMBR decoding, which assists score‑matrix completion by aligning a target metric with its distilled metrics. We evaluated it on the WMT'23 En$\leftrightarrow$De translation tasks. AC-PMBR decoding mitigated the quality degradation observed in PMBR decoding, particularly under high‑reduction settings, improved evaluation scores across metrics, and reduced approximation error by up to three times. Our study focused on distilled metrics to achieve a better cost–quality trade-off, but the framework inherently supports multi-metric ensembles, suggesting potential for multi-aspect decoding in future work.

\section*{Limitations}
\paragraph{Evaluation}
This study primarily uses BLEURT as the target metric. While applying AC-PMBR decoding to other neural metrics, such as XCOMET, would be ideal for exploring broader robustness, this short paper emphasizes the proposal of AC-PMBR decoding and its metric aggregation framework, using BLEURT as a case study. Prior work on PMBR decoding reports consistent trends and strong scores across multiple language pairs, suggesting that the underlying phenomenon is largely language-independent \citep{trabelsi-etal-2024-efficient}. Accordingly, we report results on the representative WMT’23 En$\leftrightarrow$De directions. We believe this already provides sufficient evidence to support our motivation. Similarly, although evaluating other tasks or language pairs would offer more comprehensive validation, the primary goal of this paper is to demonstrate the effectiveness of AC-PMBR decoding. Thus, we consider the current experimental scope sufficient and reserve broader extensions for future work.

\paragraph{Hardware-level Optimization}
In our experiments, we used a single GPU, but it is also possible to use two GPUs to compute the score matrices for the target metric and the distilled metric in parallel.
While further hardware-level optimizations could improve efficiency, we did not pursue them as they fall outside the core focus of this study.

\paragraph{Distillation Metric}
This study assumes the availability of a distilled metric and focuses on improving the trade-off between computational cost and translation quality under that assumption. Furthermore, the performance of AC-PMBR may vary depending on the quality of the distilled metric, as we have already demonstrated experimentally in this paper. While we use BLEURT and its existing distilled metrics as a case study, the results consistently show the effectiveness of AC-PMBR across multiple settings. Since the main focus lies in the methodological contribution, we do not explore the availability or development of better distilled metrics. Nonetheless, we expect that as research in metric distillation advances, the benefits of AC-PMBR will become even more pronounced.

\section*{Ethical Considerations}
This study fully complies with the ACL Ethics Policy and addresses all required items in the Responsible Research Checklist. All resources used in this work are publicly available and appropriately licensed, with no license-related issues. The study does not involve or generate any harmful content. While AI assistants were used for minor writing support, such as rephrasing and spell-checking, all original content was manually created by the authors. Based on the above, we confirm that this work raises no ethical concerns.

\bibliography{anthology,custom}
\newpage
\appendix
\section{Detailed Experimental Settings}
\label{sec:detailed-metrics}

\begin{table}[h]
    \centering
    \begin{tabular}{@{}lr@{}}
    \toprule
        Metrics & \#parameters \\
        \midrule
        BLEURT-20 & 579M \\
        BLEURT-20-D12 & 167M \\
        BLEURT-20-D6 & 45M \\
        BLEURT-20-D3 & 30M \\
    \bottomrule
    \end{tabular}
    \caption{Number of parameters of BLEURT-20~\citep{sellam-etal-2020-bleurt} and its distilled metrics~\citep{pu-etal-2021-learning}.}
    \label{tab:nparams:bleurt}
\end{table}

\paragraph{Datasets}
The datasets we used in our experiments and the number of sentences they contain are listed in Table~\ref{tab:datasets}, and we tuned hyperparameters only on WMT'22 En$\rightarrow$De, using WMT'23 En$\rightarrow$De and WMT'23 De$\rightarrow$En exclusively as test sets.
\begin{table}[h]
    \centering
    \tabcolsep 3pt
    \begin{tabular}{@{}lrr@{}}
        \toprule
        Dataset & En$\rightarrow$De & De$\rightarrow$En \\
        \midrule
        WMT'22 & 2,037 &  1,984\\
        WMT'23 & 557 & 549 \\
        \bottomrule
    \end{tabular}
    \caption{Number of sentences for each dataset.}
    \label{tab:datasets}
\end{table}

\paragraph{Computational costs}
The major bottleneck of AC-PMBR decoding is utility score calculation using target and distilled metrics.
In the low-reduction setting of AC-PMBR decoding, utility calculation took more than 1,000 times longer than the agreement-constrained ALS algorithm.
Therefore, \cref{alg:ac-pmbr} can be ignored from the overall cost, and the utility score calculation is dominant in the computational costs of both PMBR and AC-PMBR decoding.

We used BLEURT‑20~\cite{sellam-etal-2020-bleurt} as the utility metric, alongside its distilled metrics BLEURT‑20‑D{3,6,12}~\cite{pu-etal-2021-learning}, which shrink the parameter count from 579M to 30M, 45M, and 167M, respectively, as listed in \cref{tab:nparams:bleurt}.
Since the wall-clock computation time highly depends on the hardware environment, we evaluated the computation cost based on the time complexity.
In the PMBR decoding, the time complexity is formally defined as $\mathcal{O}(\frac{NMC}{r})$, where $C$ is the cost of utility function calls.
We fixed the number of candidates and pseudo-references, i.e., $N$ and $M$ are constant; thus, the computational costs in our settings now depend on $\mathcal{O}(\frac{C}{r})$.
Here, the cost of the utility function call $C$ is sublinearly proportional to the number of parameters in a metric model\footnote{
\url{https://github.com/google-research/bleurt/blob/master/checkpoints.md}
}.
Therefore, we defined the total computational cost of PMBR decoding $\mathrm{Cost}_\text{PMBR}$ as follows:
\begin{equation}
    \mathrm{Cost}_\text{PMBR} \coloneqq \frac{\text{\#parameters}}{r}.
\label{eq:costPMBR}
\end{equation}
Similarly, the cost of AC-PMBR decoding $\mathrm{Cost}_\text{AC-PMBR}$ is defined as follows:
\begin{equation}
    \mathrm{Cost}_\text{AC-PMBR} \coloneqq     \frac{\substack{\text{\#parameters of}\\ \text{target metric}}}{r} + \frac{\substack{\text{\#parameters of}\\ \text{distilled metric}}}{r'}.
\label{eq:costACPMBR}
\end{equation}

In all experiments, to compare the performance of PMBR and AC-PMBR decoding at a comparable cost, we set roughly the same costs for $\mathrm{Cost}_\text{PMBR}$ and $\mathrm{Cost}_\text{AC-PMBR}$, i.e., we set $\mathrm{Cost}_\text{PMBR} \approx \mathrm{Cost}_\text{AC-PMBR}$ in both high and low reduction rate settings.

\paragraph{Hyperparameter tuning}
In our AC-PMBR decoding, we tuned $\gamma \in \R_+$, which is a weight of the agreement term in \cref{eq:loss:overall}, with a fixed random seed.
As shown in Figures~\ref{fig:agreement32} and \ref{fig:agreement1024}, we varied $\gamma \in \{0.1, 0.2, \ldots 1.0\}$ in each reduction setting, i.e., low-reduction setting and high-reduction setting, and selected $\gamma=0.1$ and $\gamma=1.0$, respectively.
These tuning experiments revealed that the optimal agreement weight tends to increase with the reduction rate.
This is because, under high-reduction settings, utility scores are sparsely observed, and the information from the distilled metric becomes more beneficial for completing the score matrix.

\paragraph{Computational environment}
All experiments were conducted on a single NVIDIA RTX A6000 GPU with an Intel\regmark{} Xeon\regmark{} Gold 6426Y processor, and our method was implemented with \textsc{mbrs}~\cite{deguchi-etal-2024-mbrs}.

\onecolumn
\section{Statistical Significance Tests}
\label{sec:significance-tests}

\begin{table}[H]
  \centering
  \small
  \setlength{\tabcolsep}{6pt}
  \begin{NiceTabular}{@{}l r rr rr@{}}[colortbl-like]
    \toprule
    & & \multicolumn{2}{c}{En$\rightarrow$De} & \multicolumn{2}{c}{De$\rightarrow$En} \\
    \cmidrule(lr){3-4}\cmidrule(l){5-6}
    Decoding & & BLEU & chrF2 & BLEU & chrF2 \\
    \midrule

    \multicolumn{6}{@{}l@{}}{\rowcolor{\colorGray}\textbf{Reduction rate: Low}\quad PMBR: $r=16$, \ AC\mbox{-}PMBR: $r=32$} \\
    PMBR (baseline) & &
      19.1 $\pm$ 1.5 & 46.2 $\pm$ 1.8 &
      23.5 $\pm$ 1.7 & 50.7 $\pm$ 2.2 \\
    AC\mbox{-}PMBR & &
      19.2 $\pm$ 1.5 & \textbf{46.6} $\pm$ \textbf{1.7} &
      \textbf{23.8} $\pm$ \textbf{1.7} & \textbf{51.1} $\pm$ \textbf{2.1} \\

    \midrule

    \multicolumn{6}{@{}l@{}}{\rowcolor{\colorGray}\textbf{Reduction rate: High}\quad PMBR: $r=512$, \ AC\mbox{-}PMBR: $r=1{,}024$} \\
    PMBR (baseline) & &
      16.8 $\pm$ 1.4 & 44.1 $\pm$ 1.8 &
      21.3 $\pm$ 1.6 & 48.7 $\pm$ 2.1 \\
    AC\mbox{-}PMBR & &
      \textbf{18.1} $\pm$ \textbf{1.4} & \textbf{46.2} $\pm$ \textbf{1.6} &
      \textbf{23.0} $\pm$ \textbf{1.6} & \textbf{50.5} $\pm$ \textbf{2.0} \\

    \bottomrule
  \end{NiceTabular}

  \caption{Results of statistical significance tests on the WMT’23 En$\leftrightarrow$De translation tasks comparing AC-PMBR and PMBR decoding. 
  All scores are reported as mean $\pm$ 95\% confidence intervals. 
  Entries with $p<0.05$ and higher scores are highlighted in \textbf{bold}.}
  \label{tab:stat-sig-test}
\end{table}

\FloatBarrier

We conduct statistical significance tests using sacreBLEU~\cite{post-2018-call} for BLEU and chrF metrics, with each result based on 1{,}000 bootstrap resampling iterations drawn from the WMT’23 translation tasks. As shown in Table~\ref{tab:stat-sig-test}, the tests confirmed that the improvements achieved by AC-PMBR decoding over PMBR decoding were statistically significant ($p<0.05$) in all high-reduction settings and for chrF under the low-reduction setting, supporting the robustness of the observed gains.

\section{Detailed Derivation}
\label{sec:appendix}
In our AC-PMBR decoding, we minimize the following loss function $\mathcal{L}$:
\begin{equation}
\loss \coloneqq \lossMF(
        \mathbf{U}, \mathbf{V}; \scoreMatrixIncomplete
    ) + \lossMF(
        \mathbf{U}', \mathbf{V}'; \scoreMatrixIncomplete'
    ) + \gamma\lossAC(\mathbf{U}, \mathbf{V}, \mathbf{U}', \mathbf{V}').
\end{equation}
Thus, the rank reduced representations in the target utility, $\mathbf{u}_i$ and $\mathbf{v}_j$, are updated as follows:
\begin{align}
&\frac{\partial \loss}{\partial \mathbf{u}_i}
= -2 \sum_{
\substack{
(n, j) \in \observe{\scoreMatrixIncomplete} \\
n = i
}
} \left(O_{ij} - \mathbf{u}_i^\top \mathbf{v}_j\right) \mathbf{v}_j
+ 2\lambda\mathbf{u}_i
+ 2\gamma\left( \mathbf{u}_i - \mathbf{u}_i'\right)
\\
&\left( \sum_{
\substack{
(n, j) \in \observe{\scoreMatrixIncomplete} \\ n= i}
} \mathbf{v}_j \mathbf{v}_j^\top + (\lambda + \gamma) \mathbf{I}\right) \mathbf{u}_i
= \sum_{
\substack{
(n, j) \in \observe{\scoreMatrixIncomplete} \\ n= i}} O_{ij} \mathbf{v}_j + \gamma \mathbf{u}_i'
\\
&\mathbf{u}_i
= \left(\sum_{
\substack{
(n, j) \in \observe{\scoreMatrixIncomplete} \\ n = i}} \mathbf{v}_j \mathbf{v}_j^\top + (\lambda + \gamma) \mathbf{I}\right)^{-1}
\left(\sum_{
\substack{
(n, j) \in \observe{\scoreMatrixIncomplete} \\ n = i}} O_{ij} \mathbf{v}_j + \gamma \mathbf{u}_i'\right).
\end{align}
\begin{align}
&\frac{\partial \loss}{\partial \mathbf{v}_j}
= -2 \sum_{
\substack{
(i, m) \in \observe{\scoreMatrixIncomplete} \\ m = j}} \left(O_{ij} - \mathbf{u}_i^\top \mathbf{v}_j\right) \mathbf{u}_i
 + 2\lambda \mathbf{v}_j
+ 2\gamma \left(\mathbf{v}_j - \mathbf{v}_j'\right)
\\
&\left(\sum_{
\substack{
(i, m) \in \observe{\scoreMatrixIncomplete} \\ m = j}
} \mathbf{u}_i \mathbf{u}_i^\top + (\lambda + \gamma) \mathbf{I}\right) \mathbf{v}_j
= \sum_{
\substack{
(i, m) \in \observe{\scoreMatrixIncomplete} \\ m = j}} O_{ij} \mathbf{u}_i + \gamma \mathbf{v}_j'
\\
&\mathbf{v}_j
= \left(
\sum_{
\substack{
(i, m) \in \observe{\scoreMatrixIncomplete} \\ m = j}
} \mathbf{u}_i \mathbf{u}_i^\top + (\lambda + \gamma) \mathbf{I}\right)^{-1}
\left(\sum_{
\substack{
(i, m) \in \observe{\scoreMatrixIncomplete} \\ m = j}
} O_{ij} \mathbf{u}_i + \gamma \mathbf{v}_j'\right).
\end{align}
Likewise, the rank reduced representations in the distilled utility, $\mathbf{u}'_i$ and $\mathbf{v}'_j$, are updated as follows:
\begin{align}
&\frac{\partial \loss}{\partial \mathbf{u}_i'}
= -2 \sum_{
\substack{
(n, j) \in \observe{\scoreMatrixIncomplete'} \\ n = i}} \left(O_{ij}' - {\mathbf{u}_i'}^\top \mathbf{v}_j'\right) \mathbf{v}_j'
+ 2\lambda \mathbf{u}_i'
 + 2\gamma \left(\mathbf{u}_i' - \mathbf{u}_i\right)
\\
&\left(\sum_{
\substack{
(n, j) \in \observe{\scoreMatrixIncomplete'} \\ n = i}} \mathbf{v}_j' {\mathbf{v}_j'}^\top + (\lambda + \gamma) \mathbf{I}\right) \mathbf{u}_i'
= \sum_{
\substack{
(n, j) \in \observe{\scoreMatrixIncomplete'} \\ n = i}} O_{ij}' \mathbf{v}_j'  + \gamma\,\mathbf{u}_i,
\\
&\mathbf{u}_i'
= \left(\sum_{
\substack{
(n, j) \in \observe{\scoreMatrixIncomplete'} \\ n = i}} \mathbf{v}_j' {\mathbf{v}_j'}^\top + (\lambda + \gamma) \mathbf{I}\right)^{-1}
\left(\sum_{
\substack{
(n, j) \in \observe{\scoreMatrixIncomplete'} \\ n = i}} O_{ij}' \mathbf{v}_j' + \gamma \mathbf{u}_i\right).
\end{align}
\begin{align}
&\frac{\partial \loss}{\partial \mathbf{v}_j'}
= -2 \sum_{
\substack{
(i, m) \in \observe{\scoreMatrixIncomplete'} \\ m = j}} \left(O_{ij}' - {\mathbf{u}_i'}^\top \mathbf{v}_j'\right) \mathbf{u}_i'
+ 2\lambda\,\mathbf{v}_j'
+ 2\gamma \left(\mathbf{v}_j' - \mathbf{v}_j\right),
\\
&\left(\sum_{
\substack{
(i, m) \in \observe{\scoreMatrixIncomplete'} \\ m = j}} \mathbf{u}_i' {\mathbf{u}_i'}^\top + (\lambda + \gamma) \mathbf{I}\right) \mathbf{v}_j'
= \sum_{
\substack{
(i, m) \in \observe{\scoreMatrixIncomplete'} \\ m = j}} O_{ij}' \mathbf{u}_i' + \gamma \mathbf{v}_j,
\\
&\mathbf{v}_j'
= \left(\sum_{
\substack{
(i, m) \in \observe{\scoreMatrixIncomplete'} \\ m = j}} \mathbf{u}_i' {\mathbf{u}_i'}^\top + (\lambda + \gamma) \mathbf{I}\right)^{-1}
\left(\sum_{
\substack{
(i, m) \in \observe{\scoreMatrixIncomplete'} \\ m = j}} O_{ij}' \mathbf{u}_i' + \gamma \mathbf{v}_j\right).
\end{align}
We obtained $\mathbf{U}$ and $\mathbf{V}$ for the matrix completion using \cref{alg:ac-pmbr} with these derived update rules.

\end{document}